\pdfoutput=1
\documentclass[runningheads]{llncs}
\usepackage[T1]{fontenc}
\usepackage{graphicx}
\usepackage{mathrsfs}
\usepackage{mathtools}
\usepackage{amsfonts}
\usepackage{booktabs}
\usepackage{siunitx}
\usepackage{color, colortbl}
\usepackage{pgfplots}
\usepackage{subcaption}
\usepackage{float} 
\usepackage[export]{adjustbox}

\usepackage[font=small,labelfont=bf]{caption}
\def\Var{{\textrm{Var}}\,}
\DeclareMathOperator*{\argminA}{arg\,min} 
\definecolor{Gray}{gray}{0.9}
\pgfplotsset{compat=1.18} 
\begin{document}
\title{Benchmarking Collaborative Learning Methods Cost-Effectiveness for Prostate Segmentation}

%
\titlerunning{Benchmarking Collaborative Learning Methods Cost-Effectiveness}
%
\author{
Lucia Innocenti\inst{1,2} \and 
Michela Antonelli\inst{2} \and 
Francesco Cremonesi\inst{1} \and 
Kenaan Sarhan\inst{2} \and
Alejandro Granados\inst{2} \and
Vicky Goh\inst{2} \and
Sebastien Ourselin\inst{2} \and 
Marco Lorenzi\inst{1} 
}
\authorrunning{L. Innocenti et al.}
%
\institute{
Epione Research Group, Inria, Sophia Antipolis, France \and
King’s College London, London, UK }
%
\maketitle              
\begin{abstract}
Healthcare data is often split into medium/small-sized collections across multiple hospitals and access to it is encumbered by privacy regulations. 
This brings difficulties to use them for the development of machine learning and deep learning models, which are known to be data-hungry. 
One way to overcome this limitation is to use collaborative learning (CL) methods, which allow hospitals to work collaboratively to solve a task, without the need to explicitly share local data.

In this paper, we address a prostate segmentation problem from MRI in a collaborative scenario by comparing two different approaches: federated learning (FL) and consensus-based methods (CBM).

To the best of our knowledge, this is the first work in which CBM, such as label fusion techniques, are used to solve a problem of collaborative learning. In this setting, CBM combine predictions from locally trained models to obtain a federated strong learner with ideally improved robustness and predictive variance properties.

Our experiments show that, in the considered practical scenario, CBMs provide equal or better results than FL, while being highly cost-effective. Our results demonstrate that the consensus paradigm may represent a valid alternative to FL for typical training tasks in medical imaging.





\end{abstract}

\section{Introduction}\label{sect:introduction}
    Prostate cancer is the most frequently diagnosed cancer in men in more than half of the countries worldwide \cite{prostate_cancer}.
While accurate prostate segmentation is crucial for effective radiotherapy planning \cite{prostate_seg}, traditional manual segmentation is expensive, time-consuming, and dependent on the observer \cite{atlas_reg}.
Automated or semi-automated methods are needed for efficient and reliable prostate segmentation \cite{cad}, and deep learning is nowadays the main tool for solving the segmentation task \cite{unet}.
Hospital data are highly sensitive and are difficult to collect in data silos for centralized training. This makes their use in notoriously data-hungry deep learning systems problematic.
For this reason, collaborative learning (CL) is emerging as a powerful approach: it allows different decentralized entities to collaborate in solving a task, and researchers are exploring ways to do this by keeping the local data private \cite{fl_google}.

Federated learning (FL) \cite{future_h_fl} has gained great attention since the first apparition. FL solves a collaborative training problem in which a model is collectively optimized by different clients, each of them owning a local private dataset \cite{survey_fl}. Through different training rounds, a server orchestrates local optimization and aggregation of trained parameters across clients. Since training data is kept on the client's side, FL addresses the problems of data privacy and governance.
Nevertheless, FL still poses several challenges in real-world applications \cite{fl_problems_challenges,adv_and_probs_fl}, consisting of 1) the sensitivity of the optimization result to the heterogeneity of system and data distribution  across clients, and 2) the need for a large number of communication rounds, making communication cost a critical aspect. Moreover, from a practical perspective, FL systems are costly, since they are based on the setup and maintenance of complex computational infrastructures in hospitals, and thus require the availability of local resources and personnel \cite{realfl1,realfl2,realfl3}. 

Consensus-based methods (CBM) are a class of algorithms widely explored in machine learning, where the outputs from an ensemble of weak-predictors are aggregated to define a strong-predictor, outperforming the weak experts in terms of predictive robustness \cite{ensemble_theory}. In medical imaging, CBM are often at the core of state-of-the-art approaches for image segmentation tasks \cite{medical_ens1,medical_ens2}. 

In this paper, we propose a comparison of these two different collaborative methods. Our specific focus is on collaborative prostate segmentation applied to magnetic resonance images (MRI). Differently from FL, in CBM independent models are locally trained by each client only once and, at testing time, a strong predictor is obtained by aggregating the output of the local models.
Contrarily to FL, the setup of a CBM system in a hospital is straightforward, since no coordination in training is needed. Moreover, CBM provides data privacy and governance guarantees akin to FL, because no private information is shared during training, and model parameters are shared only once after training. Note that CBM has been coupled to FL training in previous works \cite{ensemble_fl1,ensemble_fl2,ensemble_fl3,hamer2020fedboost,guha2018oneshot,lin2020ensemble,casado2023ensemble}. Nevertheless, most of these approaches are still based on distributed optimization, and thus they require setting up the whole FL infrastructure in hospitals, while the CBM we are analyzing here overtaken this limitation.

We present in this work a thorough benchmark of these models based on a cross-silo collaborative prostate segmentation task. The contributions of this paper are the following:
\begin{itemize}
    \item We generate a distributed scenario based on natural data splits from a large collection of prostate MRI datasets currently available to the community, thus defining a realistic federated simulation.
    \item We define novel metrics to compare FL and CBM in terms of accuracy, robustness, cost-effectiveness, and utility.
    \item We apply the two CL approaches to this federated scenario and evaluate them in terms of accuracy and new-proposed metrics.
\end{itemize}

    The paper is structured as follows. In Section \ref{sect:methodology} we present the data and the learning models used for the benchmark, i.e. federated learning and consensus-based methods, and present the experiments and evaluation methods adopted in this work. Section \ref{sect:results} presents the experiment setting and results. Finally, Section \ref{sect:conclusions} discusses our findings and future perspectives.

\section{Benchmark definition}\label{sect:methodology}
    Starting from a large publicly available collection of data for prostate segmentation, we first define the federated setting by partitioning the data based on image acquisition characteristics and protocols. This allows us to obtain splits with controlled inter-center heterogeneity, thus simulating a realistic collaborative training scenario. 
We further define experiments to evaluate segmentation accuracy, cost-effectiveness, robustness to data heterogeneity, and utility for clients. Finally, we apply the differential privacy (DP) paradigm to different methods and we analyze how they respond to it.

    \subsection{Distributed Scenario}\label{subsetc:ds}
        We gathered data provided by $3$ major publicly available datasets on prostate cancer imaging analysis, and by $1$ private dataset:
\begin{itemize}
    \item \textbf{Medical Segmentation Decathlon - Prostate} \cite{decathlon} provides $32$ prostate MRIs for training.
    \item \textbf{Promise12} \cite{promise} consists of 50 training cases obtained with different scanners. Of those, $27$ cases were acquired by using an endorectal coil.
    \item \textbf{ProstateX} \cite{prostatex} contains prostate MRIs acquired by using two different scanners (Skyra and Triotim, both from Siemens). Segmentations of $194$ cases are available \cite{prostatex_masks}.
    \item \textbf{Private Hospital Dataset} (PrivateDS) is composed of 36 MRIs collected by using a Siemens Aera scanner during a project on active surveillance for prostate cancer detection. An expert radiologist produced prostate masks. This dataset is used as an independent test set. 
\end{itemize}
Datasets were split as in Table \ref{tab:nodes}, to define centers characterized by specific image acquisition properties, thus allowing to obtain heterogeneous image distributions among centers.
The common preprocessing pipeline applied to all the data comprised of flipping, cropping/padding to the same dimension, and intensity normalization. N4-bias-correction has also been applied to the data from Promise12 in N03 in order to compensate for the intensity artifacts introduced by the endorectal coil. 
\begin{table}[t]
    \centering
    \caption{Description of the different centers here considered for the distributed learning scenario, derived by partitioning the four dataset Decathlon, ProstateX, Promise12, and PrivateDS.}
    \label{tab:nodes}    
    \scalebox{0.9}{
        \begin{tabular}{r r r r c c} \toprule
        ID & \#Samples & Dataset & Subset Selection & Training & Test \\
        \midrule
        N01 & 32 & Decathlon & Full Dataset & Y & Y\\
        N02 & 23 & Promise12 & No Endorectal Coil & Y & Y\\
        N03 & 27 & Promise12 & Only Endorectal Coil & Y & Y\\
        N04 & 184 & ProstateX & Only Scanner Skyra & Y & Y\\
        N05 & 5 & ProstateX & Only Scanner Triotim & N & Y\\
        N06 & 36 & PrivateDS & Full Dataset & N & Y \\
        \bottomrule
        \end{tabular}
    }
\end{table}
    \subsection{Collaborative Learning Frameworks}\label{subsetc:frameworks}
        In our scenario we consider ${M}$ hospitals, each having a local dataset $\mathcal{D}_i = \{z_{k,i}\}_{k=1}^{N_i}$. 
Given $z$, a volumetric MRI, and a vector of parameters $\theta$, we define a segmentation problem in which a model $g$ produces binary masks  $h_{z} = g(z, \theta)$. 
Each hospital is a client indexed by $i \in [0, M]$, and the local training consists in solving the loss minimization problem, considering a loss function $f(\cdot)$.

\subsubsection{Federated learning.}
FL is a collaborative optimization problem defined by: 
\begin{equation} \label{eq:fl_loss}
       \theta_g = \argminA_{\theta}(\mathcal{L}(\theta)) \text{ s.t. } \mathcal{L}(\theta) \coloneqq \sum_{i=1}^{n} p_i\mathcal{L}_i(\theta_i).
\end{equation}
In FL, local losses are weighted by $p_i$, such that $\sum_{i=1}^{n}p_i = 1$, where the weights $p_i$ are arbitrarily set, for example, based on the local dataset size. Different strategies on how to optimize the weights have been proposed in the literature, with the aim of mitigating the impact of data heterogeneity or client drift. In this paper, we consider the following FL strategies from the state-of-the-art:
\begin{itemize}
    \item \textbf{\textsc{FedAvg}}\cite{fedavg} is the backbone of FL optimization where, at round $r$,  each client locally executes a number of stochastic gradient descent steps, and sends the partially optimized model $\theta_i^r$ to the server. The received models are weighted and averaged by the server into a global one, $\theta_g^{r+1}$, which is then sent back to the clients to initialize the next optimization round. This process is repeated for $R$ rounds until convergence. 
    \item \textbf{\textsc{FedProx}}\cite{fedprox} tackles the problem of federated optimization with data heterogeneity across clients. This approach extends \textsc{FedAvg} by introducing a proximal term to the local objective function to penalize model drift from the global optimization during local training. The proximal term is controlled by a trade-off hyperparameter, $\mu$, through the following optimization problem:
    \begin{equation} \label{eq:fedprox}
            \mathcal{L}_i(\theta)^r \coloneqq 
            \frac{1}{N_i} \sum_{k=1}^{N_i}\mathcal{L}(z_{k,i}, \theta_i^r) + \frac{\mu}{2}||\theta_i^r - \theta_g^r||^2.
    \end{equation}
 
\end{itemize}

\subsubsection{Consensus-based methods.}
With CBM, a global federated ensemble of weak predictors is composed by aggregating the outputs from the different local models. During \emph{training}, each client fully optimizes the segmentation model $g(z, \theta_i)$ on its local dataset $D_i$, by independently minimizing the local objective function $\mathcal{L}_i$. Trained local models are subsequently centralized and, for a given test image $z'$ at \emph{inference} time, the segmentation masks from all the local models are computed and aggregated by applying an ensembling strategy: 
\begin{equation}
     h_{z'} = \texttt{ensembling}(\{h_i(z')\}|_{i=1}^M) \text{ s.t. } h_i(z') = g(z',\theta_i).
\end{equation}

Among the different approaches to ensembling proposed in the literature \cite{surv_ens}, in this work we consider:
\begin{itemize}
    \item \textbf{Majority Voting} \cite{maj_vot} (\textsc{MV}) is a simple merging method that assigns to each voxel the label predicted by the majority of the local models.
    \item \textbf{\textsc{Staple}} \cite{staple} optimizes a consensus based on Expectation-Maximization (E-M) defined by the following iterative process:
    \begin{itemize}
        \item the E-step computes a probabilistic estimate of the true segmentation, that is a weighted average of each local prediction;
        \item the M-step assigns a performance level to each individual segmentation, which will be used as weights for the next E-step.
    \end{itemize}
    \item \textbf{Uncertainty-Based Ensembling} (\textsc{UBE}) is based on weighted averaging of local decisions, in which the weights represent the uncertainty of each local model on the prediction task. As uncertainty can be quantified in different ways, in this work we adopt dropout \cite{dropout_unc} to compute a measure of the global uncertainty of each local model for the segmentation of a  testing image $z$. In particular, here the uncertainty is computed as the total voxel-wise variance at inference time, defined as: $p_i = \sum_{x \in \Omega} \Var(g(z, \theta_i))[x]$, where $\Omega$ is the set of voxels in $z$, $\Var(\cdot)[x]$ is the sampling variance estimated from $S$ stochastic forward passes of the model, computed at voxel $x$.
\end{itemize}

    \subsection{Experiments details}\label{subsetc:experiments}
        The benchmark is based on four experiments, quantifying a different aspect for comparison between different strategies. 
The experiments are characterized by the same baseline model used for segmentation, which is presented below.

\textbf{Segmentation accuracy} was quantified through 5-fold cross-validation across all nodes, by testing all training strategies for each unique combination of training/testing split. The final result was obtained by averaging across all splits. 

Additionally, N05 and N06 from Table \ref{tab:nodes} were not used for training, and exclusively reserved for use as independent test set. The performance of the trained model was evaluated using the Dice Score (DSC) and a Normalised Surface Distance (NSD), following the guidelines from the Decathlon Segmentation Challenge \cite{decathlon_challenge}.

We benchmarked the following strategies. \textit{Local}: model trained only on the data from a single node, without aggregation; \textit{Centralized}: model trained on the aggregated data from all the centers; \textit{Federated}: federated training using both \textsc{FedAvg} and \textsc{FedProx} as FL strategies; \textit{Consensus}: ensembling of prediction using the CBMs strategies presented above.

\textbf{Cost-effectiveness} was investigated in terms of training and inference time and communication bandwidth \cite{costfl,costfl1,costfl2}. For estimating the bandwidth we consider the amount of data exchanged through the network during the training phase; this value depends on the model size, that in our setting is constant among all the experiments, and the number of exchanges, which is strategy-dependent.

\textbf{Model robustness} was assessed to compare FL and CBM with respect to varying data heterogeneity across clients. To this end, we evaluated the change in performance of the methods when removing N03 from the experiment. We expect a large variation in performance depending on the presence of N03, being this client the only one with images acquired with an endorectal coil, thus introducing large heterogeneity in the collaborative segmentation task. 

\textbf{Clients Utility} refers to the evaluation of how beneficial it is for an individual client to participate in a collaborative method, and which specific method would bring the most value to that client. To determine this, we consider the accuracy of different models on various test sets.

Let's consider a client labeled as $l$. We have two models: a local model denoted as $\mathcal{M}_l$, and a collaborative model denoted as $\mathcal{M}_c$. We evaluate the performance of these models on two different test sets: $\mathcal{T}_l$, which is the local test set specific to client $l$, and $\mathcal{T}_e$, which is the union of all test sets excluding $\mathcal{T}_l$.

To compare the utility of the two models, we examine two metrics:
\begin{itemize}
    \item variation in accuracy on the local test set: This is computed as the difference between the accuracy of the collaborative model on the local test set ($\text{DSC}_{\mathcal{M}_c,\mathcal{T}_l}$) and the accuracy of the local model on the same test set ($\text{DSC}_{\mathcal{M}_l,\mathcal{T}_l}$).
    \item variation in accuracy on the external test sets: This is calculated as the difference between the accuracy of the collaborative model on the combined external test sets ($\text{DSC}_{\mathcal{M}_c,\mathcal{T}_e}$) and the accuracy of the local model on the same combined external test sets ($\text{DSC}_{\mathcal{M}_l,\mathcal{T}_e}$).
\end{itemize}

By analyzing these two metrics, we can quantify the impact of using either the local or collaborative methods on both internal and external datasets. Ideally, a positive value for both metrics indicates that collaboration is beneficial for the client in all scenarios. However, it is more common to observe that collaboration improves model generalization but may affect local performance. Therefore, striking a balance between these two values is crucial.

In summary, the client's utility aims to determine the most advantageous approach for a client by comparing the accuracy variations of local and collaborative models on local and external test sets, respectively.

\textbf{Privacy mechanisms} such as differential privacy (DP) \cite{dp} have been proposed in the literature to quantify the privacy that a protocol provides and to train a model in a privacy-preserving manner. In the context of DP, the term "budget" refers to the amount of privacy protection available for the entire federated learning process, and represents the cumulative privacy loss allowed during the training phase.
This budget is typically defined as a function of $\epsilon$, where $\epsilon$ is used to control the strength of privacy guarantees for each round of federated learning updates.
Here we compared the accuracy we can obtain by spending a fixed privacy budget $\epsilon$ while protecting different collaborative methods.

A common \textbf{baseline model} was defined to obtain comparable results across strategies. We employed a 3D UNet architecture with residual connections \cite{unet}. The training was based on the optimization of the DICE Loss, by using the ADAMW optimizer for all experiments \cite{adamw}. The UNet implementation is available in the MONAI library\footnote{https://monai.io/index.html}. We fixed model hyper-parameters and maintained consistency in the amount of training, loss type, and optimizer used across all configurations. Hyperparameter search was performed by varying training parameters for all experiments (see Appendix Table $5$) and selecting those performing averagely better on the local models, obtaining a learning rate of $0.001$, a batch size of $B = 8$, and a dropout value of $0.3$. All the experiments were executed using Fed-BioMed \cite{fedbiomed}, an open-source platform that simulates the FL infrastructure. The code for running the experiments is available on the GitHub page of the author. 

The number of epochs and rounds were defined using a standard strategy \cite{flamby}, which ensured comparable numbers of training steps among local and federated training for each node. Specifically, the number of rounds $R$ for FL methods was defined as follows: $R = E \cdot N_T/M/B/s$, where $E$ is the number of epochs required to train the model locally, $s = 20$ is the fixed number of local SGD steps, and $N_T$ is the total number of samples in the training set.
    
\section{Results}\label{sect:results}
    
\begin{figure}[t] 
\includegraphics[page=1,trim={0 1.2cm 0 7cm},clip,width=\linewidth]{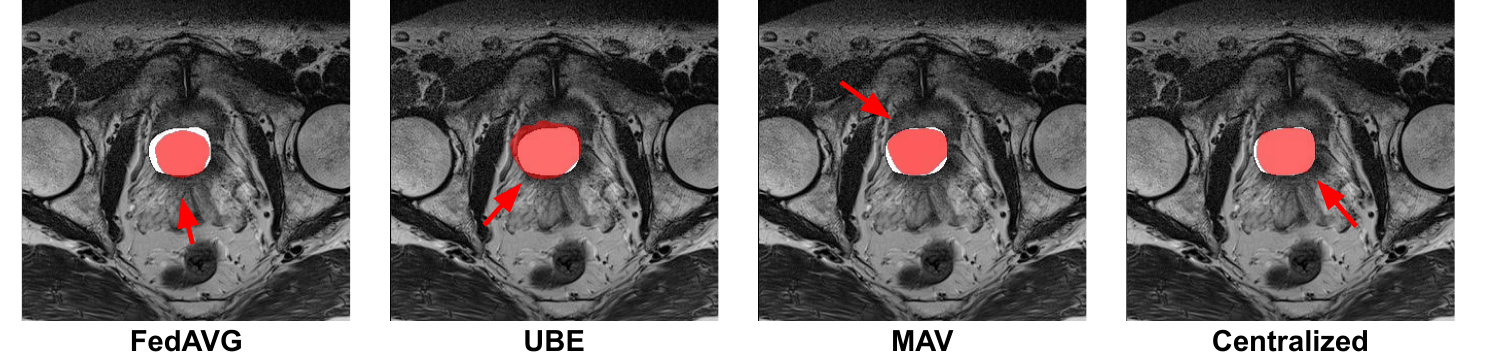}
\caption{A representation of the segmentation task on a sample image using different strategies. In white, the ground truth; in red, the segmentation provided by each training approach.} \label{fig:results}
\end{figure}

\textbf{Segmentation Accuracy.}
Table \ref{tab:results} presents the average DSC among the 5-Fold evaluations obtained with the different collaborative learning strategies, while an illustrative example of the results on a sample image is available in Figure \ref{fig:results}. The best results are indicated in \textbf{bold}. Similar results are obtained with the NSD metric and can be found in Appendix Table $1$ and Table $2$. Details about standard deviation among the K runs can be found in Appendix Table~$3$.

Overall, CBM obtain better or at least comparable results than FL: the last row in Table~\ref{tab:results} shows that \textsc{UBE} is on average the best-performing method, but all the CBM provide very similar results. In general, distributed methods highly outperform local methods, which fail to generalize.

\newcolumntype{g}{>{\columncolor{Gray}}c}
\newcolumntype{t}{>{\columncolor{Gray}}r}
\begin{table}[t]
    \centering
    \caption{Comparison of the 5-fold DSC obtained in the segmentation task by different training strategies.}
    \label{tab:results}
    \resizebox{\textwidth}{!}{%
    \begin{tabular}{@{}c||ttttc||ttccc@{}}\toprule
     &
    \multicolumn{4}{g}{\textbf{Local}} &
    \textbf{Centralized} &
    \multicolumn{2}{g}{\textbf{Federated}} &
    \multicolumn{3}{c}{\textbf{Consensus}} \\
    & N01 & N02 & N03 & N04 & &\textsc{FedAvg} &\textsc{FedProx}&\textsc{UBE}&\textsc{Staple}&\textsc{MV}  \\   
    \midrule     
     N01-test &0.86 & 0.64 & 0.49 & 0.44 & 0.92 & 0.85 & 0.70 & \textbf{0.89} & 0.83 & 0.84\\
     N02-test &0.80 & 0.69 & 0.66 & 0.73 & 0.90 & 0.82 & 0.75 & 0.85 & \textbf{0.87} & \textbf{0.87}\\ 
     N03-test &0.64 & 0.72 & 0.75 & 0.44 & 0.83 & 0.70 & 0.75 & 0.73 & 0.75 & \textbf{0.76}\\
     N04-test &0.79 & 0.66 & 0.62 & 0.88 & 0.91 & \textbf{0.88} & 0.84 & 0.87 & 0.86 & 0.86\\     
     N05 &0.57 & 0.68 & 0.71 & 0.73 & 0.77 & 0.71 & 0.67 & \textbf{0.72} & 0.68 & 0.68\\
     N06 &0.75 & 0.63 & 0.61 & 0.75 & 0.83 & \textbf{0.82} & 0.80 & 0.80 & \textbf{0.82} &\textbf{ 0.82}\\
     \midrule
     Average &0.73 & 0.67 & 0.64 & 0.66 & 0.86 & 0.80 & 0.75 & \textbf{0.81} & 0.80 & 0.80\\
    \bottomrule
    \end{tabular}
    }
\end{table}

\textbf{Cost-Effectiveness.} We consider the total training time for FL and the longest time for local training across clients for CBM. Federated training is roughly three times longer than CBM training ($\sim 2$ hours vs $\sim 30$ minutes). Among the CBM methods, UME is associated with the largest testing time, having to perform many inferences to estimate the uncertainty map. MAV is the most efficient and takes two times longer than the average FL (though still in the order of seconds). However, we note that testing time is a magnitude lower than training time, making its impact irrelevant in a real case application. The amount of exchanged data for FL is equal to $2 \cdot M \cdot m_s \cdot R$, where $R$ is the number of rounds and $m_s$ is the model size. For CBM, is only $M \cdot m_s$, resulting in a difference of $C \cdot m_s \cdot (2\cdot R - 1)$. Considering the UNet used in the experiment, $m_s = 30MB$, the difference between FL and CBM is roughly of $9.25$ GBytes.

\begin{table}[t]
    \centering
    \caption{Comparison of costs of different training strategies in terms of training and inference time and training bandwidth.}
    \label{tab:costeffect}
    \resizebox{\textwidth}{!}{%
    \begin{tabular}{@{}r||ttttc||ttccc@{}}\toprule
     &
    \multicolumn{4}{g}{\textbf{Local}} &
    \textbf{Centralized} &
    \multicolumn{2}{g}{\textbf{Federated}} &
    \multicolumn{3}{c}{\textbf{Consensus}} \\
    & N01 & N02 & N03 & N04 & &\textsc{FedAvg} &\textsc{FedProx}&\textsc{UBE}&\textsc{Staple}&\textsc{MV}  \\   
    \midrule     
    Train. time (min) & 22 & 35 & 38 & 36 & 421 & 116 & 116 & \textbf{38} & \textbf{38} & \textbf{38} \\
    Inf. time (sec) & 0.4 & 0.4 & 0.4 & 0.4 & 0.4 & \textbf{0.3} & \textbf{0.3} & 16.3 & 3.7 & 0.9 \\
    Train. Bandwidth (\SI{}{MB}) & 30 & 30 & 30 & 30 & 0 & 9600 & 9600 & \textbf{120} & \textbf{120} & \textbf{120} \\
    \bottomrule
    \end{tabular}
    }
\end{table}

\textbf{Model Robustness.} The performance of local models reported in Table \ref{tab:results} (panel "Local") allows to appreciate the heterogeneity across clients. 
As expected, N03 emerges as the client with the highest heterogeneity from this analysis, given the drop in testing performance of the models locally trained on the other clients. 
As shown in Appendix Table~$4$, CBM leads to an average absolute DSC variation of $1.7 \%$, $2.4\%$, and $2.7\%$, for respectively UBE, MV, and Staple, as compared to the $3.1\%$ and $5.7\%$ DSC change respectively associated with \textsc{FedAvg} and \textsc{FedProx}. A graphical representation of this property is available in Appendix Figure~$1$.
This result denotes the improved robustness of CBM to clients' heterogeneity. The overall results obtained after removing N03 are compatible with those shown in Table~\ref{tab:results}, and confirm the positive performances of CBM as compared to FL.

\textbf{Clients Utility.} Figure~\ref{tab:utility} presents a comparison of the utility of different collaborative methods for the four clients in the experiment. 
\definecolor{chartred}{RGB}{225,113,113}
\definecolor{chartblue}{RGB}{140,223,214}
\begin{figure}[H]
  \centering
  \begin{subfigure}[b]{0.49\linewidth}
  \resizebox{\textwidth}{!}{%
    \begin{tikzpicture}
      \begin{axis}[
        ybar,
        ymin=-20,
        ymax=10,
        symbolic x coords={FedAVG, FedProx, UBE, STAPLE, MV},
        xtick=data,
        ticklabel style={font=\small},
        legend style={at={(0.5,-0.15)},
          anchor=north,legend columns=-1},
        ymajorgrids=true,
        grid style={dashed, gray!50},
        ytick={-20, -15, -10, -5, 0, 5, 10},
        yticklabels={-20\%, -15\%, -10\%, -5\%, 0\%, 5\%, 10\%},
      ]

      \addplot[fill=chartblue] coordinates {(FedAVG, 8.1) (FedProx, 5.5) (UBE, 8.4) (STAPLE, 8.8) (MV, 8.9)};
      \addplot[fill=chartred] coordinates {(FedAVG, -0.6) (FedProx, -15.8) (UBE, -1.8) (STAPLE, -2.6) (MV, -2.4)};
      \end{axis}
    \end{tikzpicture}
    }
   \caption{Analysis on utility for client N01}
  \end{subfigure}
  \hfill
  \begin{subfigure}[b]{0.49\linewidth}
  \resizebox{\textwidth}{!}{%
      \begin{tikzpicture}
        \begin{axis}[
          ybar,
          ymin=0,
          ymax=20,
          symbolic x coords={FedAVG, FedProx, UBE, STAPLE, MV},
          xtick=data,
          ticklabel style={font=\small},
          legend style={at={(0.5,-0.15)},
            anchor=north,legend columns=-1},
          ymajorgrids=true,
          grid style={dashed, gray!50},
          ytick={0, 5, 10, 15, 20},
          yticklabels={0\%, 5\%, 10\%, 15\%, 20\%},
        ]

        \addplot[fill=chartblue] coordinates {(FedAVG, 13) (FedProx, 8.9) (UBE, 13.5) (STAPLE, 12.6) (MV, 12.7)};
        \addplot[fill=chartred] coordinates {(FedAVG, 13.2) (FedProx, 5.8) (UBE, 16) (STAPLE, 17.6) (MV, 17.6)};
        \end{axis}
      \end{tikzpicture}
    }
    \caption{Analysis on utility for client N02}
  \end{subfigure}
  
  \begin{subfigure}[b]{0.49\linewidth}
  \resizebox{\textwidth}{!}{%
    \begin{tikzpicture}
        \begin{axis}[
          ybar,
          ymin=-5,
          ymax=30,
          symbolic x coords={FedAVG, FedProx, UBE, STAPLE, MV},
          xtick=data,
          ticklabel style={font=\small},
          legend style={at={(0.5,-0.15)},
            anchor=north,legend columns=-1},
          ymajorgrids=true,
          grid style={dashed, gray!50},
        ytick={-5,0,5,10,15,20, 25, 30},
        yticklabels={-5\%,0\%,5\%,10\%,15\%,20\%, 25\%, 30\%},
        ]

        \addplot[fill=chartblue] coordinates {(FedAVG, 25) (FedProx, 18.6) (UBE, 25.8) (STAPLE, 24.6) (MV, 24.7)};
        \addplot[fill=chartred] coordinates {(FedAVG, 5) (FedProx, -0.4) (UBE, -2.4) (STAPLE, 0.4) (MV, 0.6)};
        \end{axis}
      \end{tikzpicture}
    }
    \caption{Analysis on utility for client N03}
  \end{subfigure}%
  \hfill
  \begin{subfigure}[b]{0.49\linewidth}
  \resizebox{\textwidth}{!}{%
    \begin{tikzpicture}
      \begin{axis}[
        ybar,
        ymin=-5,
        ymax=20,
        symbolic x coords={FedAVG, FedProx, UBE, STAPLE, MV},
        xtick=data,
        ticklabel style={font=\small},
        legend style={at={(0.5,-0.15)},
          anchor=north,legend columns=-1},
        ymajorgrids=true,
        grid style={dashed, gray!50},
        ytick={-5,0,5,10,15,20},
        yticklabels={-5\%,0\%,5\%,10\%,15\%,20\%},
      ]

      \addplot[fill=chartblue] coordinates {(FedAVG, 16.5) (FedProx, 11.8) (UBE, 17.9) (STAPLE, 17.4) (MV, 17.6)};
      \addplot[fill=chartred] coordinates {(FedAVG, 0) (FedProx, -4.6) (UBE, -1.8) (STAPLE, -2.6) (MV, -2.4)};
      \end{axis}
    \end{tikzpicture}
    }
    \caption{Analysis on utility for client N04}
  \end{subfigure}
  
  \caption{The chart shows the utility of collaborative methods with respect to local models when used on the local test sets (red bar) or external test sets (blue bar) for each client indicated in the sub-captions. Each histogram corresponds to a different client. 
  For all clients, collaborative methods improved generalization by a difference of up to $25\%$, while decreasing local performance by at most $15\%$ and in some cases even improving it. A significant degree of heterogeneity can be observed in the impact on generalization and local performance among different test sets as well as different methods. }
  \label{tab:utility} 
\end{figure}
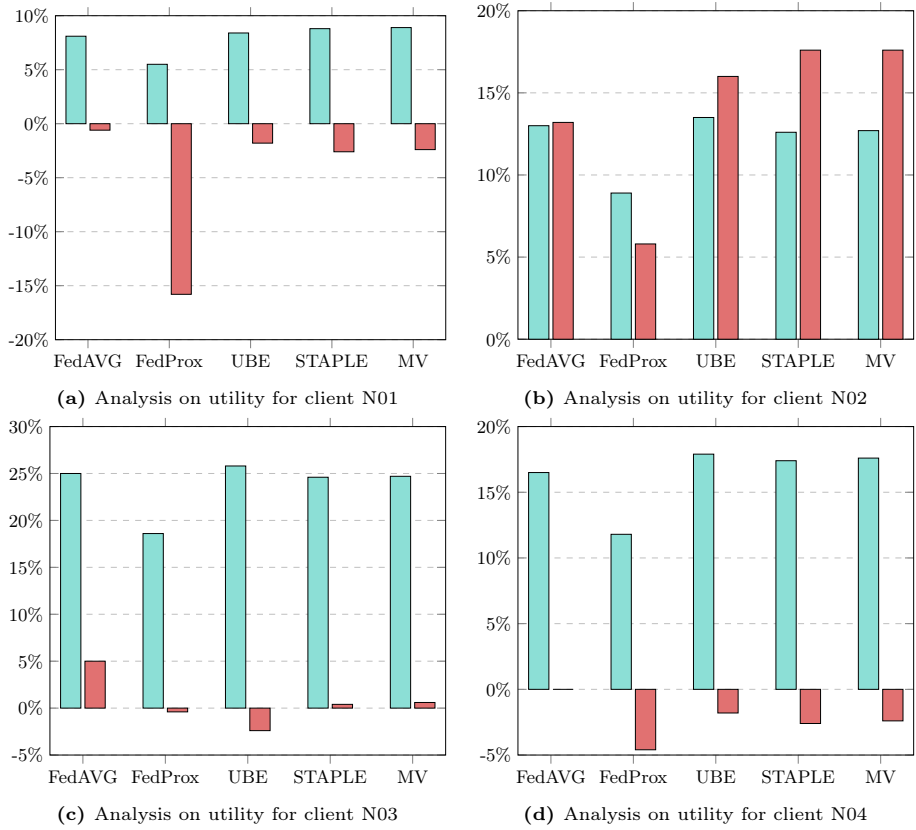
For all clients and all methods, collaborative methods lead to improvements in model generalization when evaluated on external test sets. This implies that collaborating with other clients helps to enhance the overall performance of the models on unseen data. Additionally, it is worth noting that even for small clients like N02, collaborative methods also result in improved local performance. This suggests that even clients with limited local data can benefit from participating in the collaboration.
Surprisingly, even the largest client, N04, still experiences advantages by joining the collaboration. This indicates that size alone does not diminish the benefits of collaborative methods and that even clients with substantial local datasets can gain value from collaboration.
Overall, in this particular experiment, the performance of CBM is comparable to that of FL. However, UBE method consistently demonstrates the most substantial improvements across various metrics, making it the preferred choice among the collaborative methods evaluated.

\textbf{Privacy mechanisms.} The privacy analysis is performed in the framework of R\'enyi Differential Privacy (RDP)~\cite{renyi_dp}, a relaxation of the classical definition~\cite{dwork2006calibrating} allowing a convenient way to keep track of the cumulative privacy loss. This allows us to quantify the privacy budget $\epsilon$ corresponding to SGD optimization with parameters defined as for the baseline model of Section~\ref{subsetc:experiments}. 
\definecolor{chartred}{RGB}{225,113,113}
\definecolor{chartblue}{RGB}{140,223,214}
\begin{figure}[H]
  \centering
  \begin{tikzpicture}[scale=0.8]
    \begin{axis}[
      xlabel={$\epsilon$},
      ylabel={$DSC$ score},
      legend style={at={(0.5,-0.2)}, anchor=north},
      xmin=0.5, xmax=5.5,
      ymin=0, ymax=1,
      xtick={0.5,1,1.5,2,2.5,3,3.5,4,4.5,5,5.5},
      ytick={0,0.1,0.2,0.3,0.4,0.5,0.6,0.7,0.8,0.9,1},
      grid=both,
      grid style={dashed,line width=0.2pt, draw=gray!50},
    ]
    
    \addplot[color=chartblue,mark=none,smooth,line width=1.5pt] coordinates {
      (0.61,0.27)
      (1.02,0.36)
      (1.43,0.37)
      (1.81,0.52)
      (2.05,0.78)
      (2.46,0.79)
      (2.66,0.80)
      (2.89,0.81)
      (3.12,0.77)
      (3.45,0.82)
      (3.69,0.83)
      (3.89,0.83)
      (4.10,0.82)
      (4.35,0.82)
      (4.55,0.82)
      (4.71,0.82)
      (4.99,0.84)
      (5.20,0.84)
    };
    
    \addplot[color=chartred,mark=none,smooth,line width=1.5pt] coordinates {
      (0.61,0.14)
      (1.02,0.18)
      (1.43,0.59)
      (1.81,0.47)
      (2.05,0.58)
      (2.46,0.65)
      (2.66,0.67)
      (2.89,0.64)
      (3.12,0.71)
      (3.45,0.68)
      (3.69,0.71)
      (3.89,0.76)
      (4.10,0.75)
      (4.35,0.71)
      (4.55,0.74)
      (4.71,0.77)
      (4.99,0.75)
      (5.20,0.74)
    };
    
    \legend{CBM (MV), FL (FedAvg)}
    
    \end{axis}
  \end{tikzpicture}
  \caption{The chart compares the accuracy reached by different methods when spending a privacy budget $\epsilon$ for differential privacy. The two compared methods are majority voting (MV) for CBM and federated averaging (FedAvg) for FL. CBM obtain on average better performances when $\epsilon$ is fixed, and it already reaches the plateau with $\epsilon \approx 3$.}
  \label{fig:dp_comparison}
\end{figure}
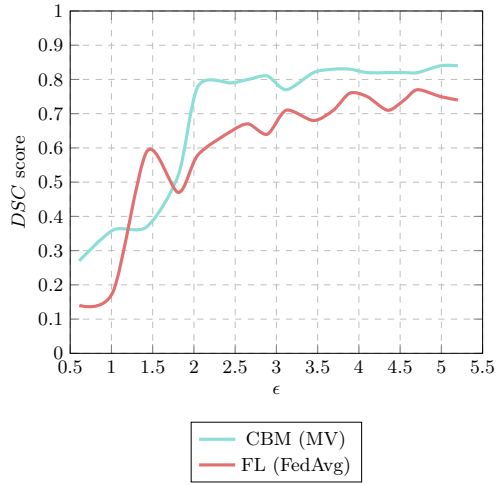
Following~\cite{dl_with_dp}, the DP Gaussian mechanism was defined with noise $\sigma = 4$. One can show that the privacy budget for obtaining the results presented in Table~\ref{tab:results} is $\epsilon_{CBM} = 5.2$ in the CBM scenario, and $\epsilon_{FL} = 7.9$ in the federated one, denoting the lower privacy cost of CBM. 
CBM is also characterized by a lower privacy cost in relation to our chosen performance metric: DSC. We compared how DSC on unseen data evolved for the ensembling method MV and the FedAVG aggregation strategy when the privacy budget $\epsilon$ varied between $0.5$ and $5.5$. Figure~\ref{fig:dp_comparison} shows that the CMB method achieved a higher DSC than FL with a lower privacy budget: CBM reached a plateau at roughly $\epsilon = 3$ while FL reached a plateau only after $\epsilon > 4$, and at a lower DSC value.

\section{Conclusions}\label{sect:conclusions}
    In this paper, we proposed a realistic benchmark for collaborative learning methods for prostate segmentation.
To this end, we used a collection of large public and private prostate MRI datasets to simulate a realistic distributed scenario across hospitals and we defined experiments and metrics to compare local training with different collaborative learning methods, namely FL and CBM, in terms of performances, cost-effectiveness, robustness and privacy of the models.
For the considered scenario of cross-silo federated prostate segmentation, our results show that CBM represent a reliable alternative to FL in terms of performances, while being highly competitive in terms of robustness, and superior in cost-effectiveness when considering the practical implementation and required resources. Indeed, CBM avoid synchronization of training across hospitals, while the setup of an FL infrastructure is costly and time-consuming, and often prohibitive for typical hospital applications. 

By simply sharing locally trained models and applying CBM to local predictions, we can rely on established theory from the state-of-the-art of multi-atlas segmentation to obtain competitive results at much less cost, as CBM avoid synchronization of training across hospitals. 
 
Our preliminary results on  privacy-preserving methods based on differential privacy show that CBM can guarantee a stronger level of privacy protection.

Moreover, secure aggregation techniques could be used at inference time for CBM in order to avoid sharing the whole model, adding another privacy layer to the framework. Other FL schemes could be included in our benchmark, such as \textsc{SCAFFOLD} \cite{scaffold} or \textsc{FedOpt} \cite{fedopt}, to better account for heterogeneity. Nevertheless, given previous benchmark results on similar medical imaging tasks \cite{flamby}, we do not expect a substantial change in the overall message of this study, especially concerning the comparison of cost-effectiveness between FL and CBM paradigms. Different consensus strategies could be implemented in the future, for example, to account for voxel-wise uncertainty across local models. 
The benchmark here proposed focuses on a cross-silo setup, typical of FL applications in hospitals proposed so far. Future investigations could extend our study to include a larger number of clients, thus allowing to better exploit the robustness guarantees associated with consensus strategies.

\bibliographystyle{unsrt}
\bibliography{biblio}

\section{Supplementary Material}\label{sect:supp}


\begin{table}[H]
    \centering
    \caption{Comparison of the NSD obtained in the segmentation task by different training strategies on the different test sets. Training strategies are grouped by approach.}
    \label{tab:nds_results}
    \resizebox{\textwidth}{!}{%
    \begin{tabular}{@{}cttttcttccc@{}} \midrule
     &
    \multicolumn{4}{g}{\textbf{Local}} &
    \textbf{Centralized} &
    \multicolumn{2}{g}{\textbf{Federated}} &
    \multicolumn{3}{c}{\textbf{Consensus}} \\
    & N01 & N02 & N03 & N04 & &FedAvg &FedProx&UBE&Staple&MV  \\   
    \toprule     
     N01-test &0.82 & 0.62 & 0.22 & 0.45 & 0.91 & 0.86 & 0.66 & \textbf{0.89} & 0.83 & 0.83\\
     N02-test &0.64 & 0.63 & 0.37 & 0.74 & 0.91 & 0.84 & 0.71 &\textbf{ 0.85} & 0.80 & 0.80\\ 
     N03-test &0.41 & 0.54 & 0.73 & 0.36 & 0.78 & 0.5 & \textbf{0.64} & \textbf{0.64} & 0.58 & 0.54 \\
     N04-test &0.67 & 0.65 & 0.27 & 0.92 & 0.94 & \textbf{0.89} & 0.76 & 0.87 & 0.83 & 0.83\\     
     N05 &0.37 & 0.46 & 0.18 & 0.63 & 0.62 & 0.64 & 0.57 & \textbf{0.74} & 0.61 & 0.60\\
     N06 &0.39 & 0.52 & 0.26 & 0.68 & 0.75 & 0.66 & 0.55 & \textbf{0.71} & 0.60 & 0.58\\
     \midrule
     Average &0.55 & 0.57 & 0.34 & 0.63 & 0.82 & 0.73 & 0.65 & \textbf{0.78} & 0.71 & 0.70\\
    \bottomrule
    \end{tabular}
    }
\end{table}


\begin{table}[H]
    \centering
    \caption{Comparison of the deviation standard in the NSD obtained in the segmentation task by different training strategies on the different test sets. Training strategies are grouped by approach.}
    \label{tab:nds_ds}
    \resizebox{\textwidth}{!}{%
    \begin{tabular}{@{}cttttcttccc@{}} \midrule
     &
    \multicolumn{4}{g}{\textbf{Local}} &
    \textbf{Centralized} &
    \multicolumn{2}{g}{\textbf{Federated}} &
    \multicolumn{3}{c}{\textbf{Consensus}} \\
    & N01 & N02 & N03 & N04 & &FedAvg &FedProx&UBE&Staple&MV  \\   
    \toprule     
     N01-test &0.09 & 0.32 & 0.31 & 0.14 & 0.06 & 0.30 & 0.07 & 0.04 & 0.05 & 0.05\\
     N02-test &0.15 & 0.32 & 0.38 & 0.10 & 0.09 & 0.38 & 0.07 & 0.06 & 0.09 & 0.09\\ 
     N03-test &0.28 & 0.33 & 0.07 & 0.19 & 0.34 & 0.23 & 0.24 & 0.23 & 0.33 & 0.28\\
     N04-test &0.06 & 0.32 & 0.38 & 0.02 & 0.13 & 0.39 & 0.04 & 0.03 & 0.03 & 0.03\\     
     N05 &0.21 & 0.25 & 0.38 & 0.16 & 0.09 & 0.33 & 0.15 & 0.05 & 0.21 & 0.21\\
     N06 &0.17 & 0.20 & 0.28 & 0.03 & 0.14 & 0.28 & 0.08 & 0.01 & 0.03 & 0.04\\
     \midrule
     Average &0.16 & 0.29 & 0.30 & 0.11 & 0.14 & 0.32 & 0.11 & 0.07 & 0.12 & 0.12\\
    \bottomrule
    \end{tabular}
    }
\end{table}


\begin{table}[H]
    \centering
    \caption{Comparison of the deviation standard in the DSC obtained in the segmentation task by different training strategies on the different test sets. Training strategies are grouped by approach.}
    \label{tab:dsc_ds}
    \resizebox{\textwidth}{!}{%
    \begin{tabular}{@{}cttttcttccc@{}} \midrule
     &
    \multicolumn{4}{g}{\textbf{Local}} &
    \textbf{Centralized} &
    \multicolumn{2}{g}{\textbf{Federated}} &
    \multicolumn{3}{c}{\textbf{Consensus}} \\
    & N01 & N02 & N03 & N04 & &FedAvg &FedProx&UBE&Staple&MV  \\   
    \toprule     
     N01-test &1.0E-01 & 3.3E-01 & 2.6E-01 & 1.9E-01 & 1.9E-02 & 6.4E-02 & 1.8E-01 & 3.4E-02 & 9.0E-02 & 8.3E-02\\
     N02-test &6.1E-02 & 3.4E-01 & 1.5E-01 & 1.1E-01 & 2.4E-02 & 7.0E-02 & 1.3E-01 & 3.7E-02 & 2.9E-02 & 2.9E-02\\ 
     N03-test &1.9E-01 & 1.7E-01 & 5.1E-02 & 1.4E-01 & 1.1E-01 & 1.6E-01 & 1.2E-01 & 1.3E-01 & 1.5E-01 & 1.5E-01\\
     N04-test &7.0E-04 & 1.0E-01 & 3.1E-02 & 1.3E-04 & 3.8E-04 & 4.4E-03 & 2.3E-02 & 2.3E-02 & 1.8E-02 & 2.5E-04\\     
     N05 &3.6E-02 & 3.3E-01 & 1.9E-01 & 6.7E-02 & 3.6E-02 & 3.9E-02 & 1.1E-01 & 3.3E-02 & 1.3E-02 & 1.1E-02\\
     N06 &1.8E-02 & 2.7E-02 & 2.6E-02 & 2.1E-02 & 6.0E-02 & 1.5E-02 & 1.6E-02 & 1.3E-02 & 2.1E-02 & 1.8E-02\\
     \midrule
     Average &5.1E-02 & 2.8E-01 & 1.6E-01 & 9.3E-02 & 2.0E-02 & 4.4E-02 & 1.1E-01 & 4.4E-02 & 3.7E-02 & 3.1E-02\\
    \bottomrule
    \end{tabular}
    }
\end{table}

\begin{table}[H]
    \centering
    \caption{Absolute differences in DSC score, among all the test sets,  between the training with and without N03. The lower the difference, the higher the robustness to data heterogeneity for the strategy.}
    \label{tab:robustness}
    \begin{tabular}{@{}crrrrr@{}} \toprule
    & \textsc{FedAvg} & \textsc{FedProx} & \textsc{UBE} & \textsc{Staple} & \textsc{MV} \\
    \midrule
    N01-test & 6.0 \% & \textbf{4.2}\% & 0.6 \% & 7.8 \% & 6.8 \% \\
    N02-test & \textbf{4.6} \% & 7.2\% & 5.0 \% & 6.4 \% & 6.2 \% \\
    N03-test & \textbf{0.6} \% & 1.8\% & 1.4 \% & 1.0 \% & 0.8 \% \\
    N04-test & 2.2 \% & 5.4\% & 0.8 \% & \textbf{0.2} \% & 0.4 \% \\
    N05 & 2.0 \% & 8.8\% & 2.0 \% & 0.4 \% & \textbf{0.0} \% \\
    N06 & 3.2 \% & 4.6\% & 0.4 \% & 0.6 \% & \textbf{0.2} \% \\
    \midrule
    Average & 3.1 \% & 4.3\% & \textbf{1.7} \% & 2.7 \% & 2.4 \% \\
    \bottomrule
    \end{tabular}
\end{table}
\begin{figure}[H] 
\includegraphics[page=3,trim={0 6.5cm 0 1cm},clip,width=\linewidth]{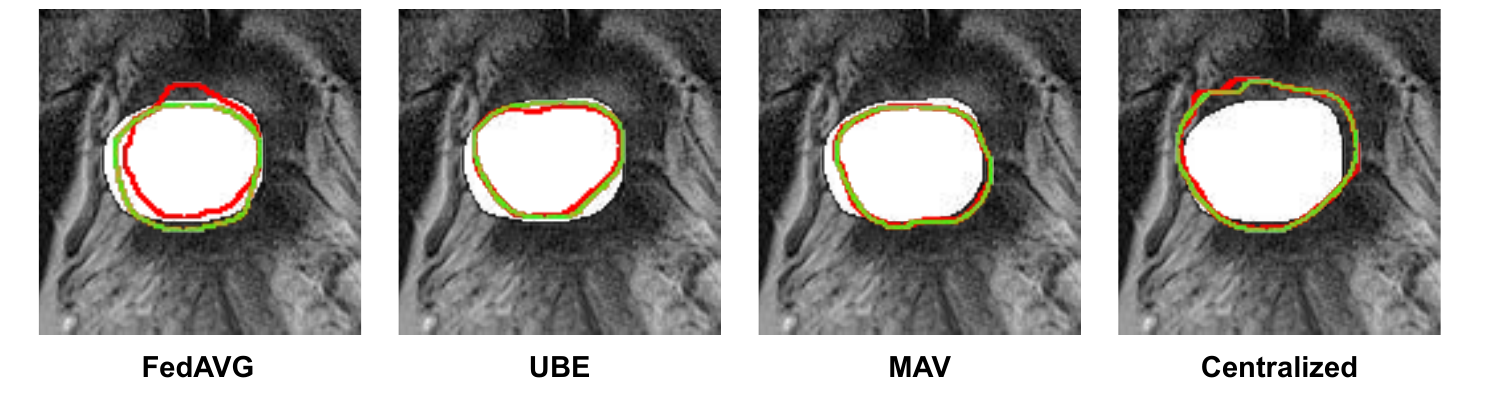}
\caption{A representation of the segmentation task on the same sample image using different strategies, comparing the model trained with, or without, client N03. In white, the ground truth; the red the contour obtained by the model trained with N03; in blue, the opposite.
 } \label{fig:rob}
\end{figure}


\begin{table}[H]
    \centering
    \caption{Hyperparameters and respective values explored during the tuning phase. Selected value in \textbf{bold}. The selection of dropout value was driven by the need to use it for the \textsc{UBE} method. Finally, we set the number of epochs for local training $E = \frac{1}{M}\sum_{i=1}^M{E_i}$, where $E_i$ is the number of epochs computed by each local node while taking $K$ gradient steps}
    \label{tab:params}
    \begin{tabular}{@{}rr@{}} \toprule
    Parameter & Values \\
    \midrule
    Learning Rate & 0.0001; \textbf{0.001}; 0.01; 0.1; 1 \\
    Batch Size & 4, \textbf{8}, 16 \\
    Dropout & 0.1,\textbf{0.3},0.5 \\
    Local Steps & 10, 15, \textbf{20}, 25\\
    K & 300, 400, \textbf{450}, 500\\
    \bottomrule
    \end{tabular}
\end{table}

\end{document}